\documentclass[letterpaper, 10 pt, conference]{ieeeconf}  
\IEEEoverridecommandlockouts                              
\usepackage{amsfonts,amssymb}
\usepackage{multirow}
\usepackage{bbm}
\usepackage{url}
\usepackage{flushend}
\usepackage{amsmath}
\usepackage{algorithm}
\usepackage{algpseudocode}
\usepackage{amsfonts}
\usepackage{graphics} 
\usepackage{graphicx}
\usepackage{array}
\usepackage{amsmath}
\usepackage{hyperref}
\usepackage{mathtools}
\usepackage{cite}
\usepackage{textcomp}
\usepackage{cleveref}
\usepackage{times}
\usepackage{epsfig}
\usepackage{gensymb}
\usepackage{multirow}
\usepackage{caption}
\usepackage{pifont}
\usepackage{subfigure}
\usepackage{url}
\usepackage{tikz}
\usepackage{cleveref}
\usepackage{mathtools,xparse}
\usepackage[super]{nth}
\DeclareMathOperator*{\argminB}{argmin}   

\title{\LARGE \bf EAO-SLAM: Monocular Semi-Dense Object SLAM \\ Based on Ensemble Data Association}

\author{Yanmin Wu$^{1}$, Yunzhou Zhang$^{1,2}$, Delong Zhu$^{3}$, Yonghui Feng$^{2}$, Sonya Coleman$^{4}$ and Dermot Kerr$^{4}$
\thanks{This work was supported by National Natural Science Foundation of China (No. 61973066,61471110) , Equipment Pre-research Fundation(61403120111), the Fundation of Key Laboratory of  Aerospace System Simulation(6142002301), the Fundation of Key Laboratory of Equipment Reliability(61420030302), Natural Science Foundation of Liaoning (No.20180520040), and Fundamental Research Funds for the Central Universities(N172608005, N182608004).}%
\thanks{$^{1}$Yanmin Wu is with Faculty of Robot Science and Engineering, Northeastern University, Shenyang, China.}%
\thanks{$^{2}$Yunzhou Zhang and Yonghui Feng are with College of Information Science and Engineering, Northeastern University, Shenyang 110819, China (Corresponding author: Yunzhou Zhang, Email: {\tt\small zhangyunzhou@mail.neu.edu.cn}).}%
\thanks{$^{3}$Delong Zhu is with the Department of Electronic Engineering, The Chinese University of Hong Kong, Shatin, N.T., Hong Kong SAR, China.}
\thanks{$^{4}$Sonya Coleman and Dermot Kerr are with School of Computing and Intelligent Systems,Ulster University, N. Ireland, UK.}}

\begin{document}
\maketitle

\begin{abstract}
Object-level data association and pose estimation play a fundamental role in semantic SLAM, which remain unsolved due to the lack of robust and accurate algorithms. In this work, we propose an ensemble data associate strategy for integrating the parametric and nonparametric statistic tests. By exploiting the nature of different statistics, our method can effectively aggregate the information of different measurements, and thus significantly improve the robustness and accuracy of data association. We then present an accurate object pose estimation framework, in which an outliers-robust centroid and scale estimation algorithm and an object pose initialization algorithm are developed to help improve the optimality of pose estimation results. 
Furthermore, we build a SLAM system that can generate semi-dense or lightweight object-oriented maps with a monocular camera.
Extensive experiments are conducted on three publicly available datasets and a real scenario. The results show that our approach significantly outperforms state-of-the-art techniques in accuracy and robustness. The source code is available on \url{https://github.com/yanmin-wu/EAO-SLAM}. 
\end{abstract}

\section{Introduction}
Conventional visual SLAM systems have achieved significant success in robot localization and mapping tasks. More efforts in recent years are evolved in making SLAM serve for robot navigation, object manipulation, and environment representation. Semantic SLAM is a promising technique for enabling such applications and receives much attention from the community~\cite{wang2020tartanair}. 
In addition to the conventional functions, semantic SLAM also focuses on a detailed expression of the environment, e.g., labeling map elements or objects of interests, to support different high-level applications. 

\begin{figure}[t]
	\vspace{2mm}
	\centering
	\includegraphics[width=0.44\textwidth]{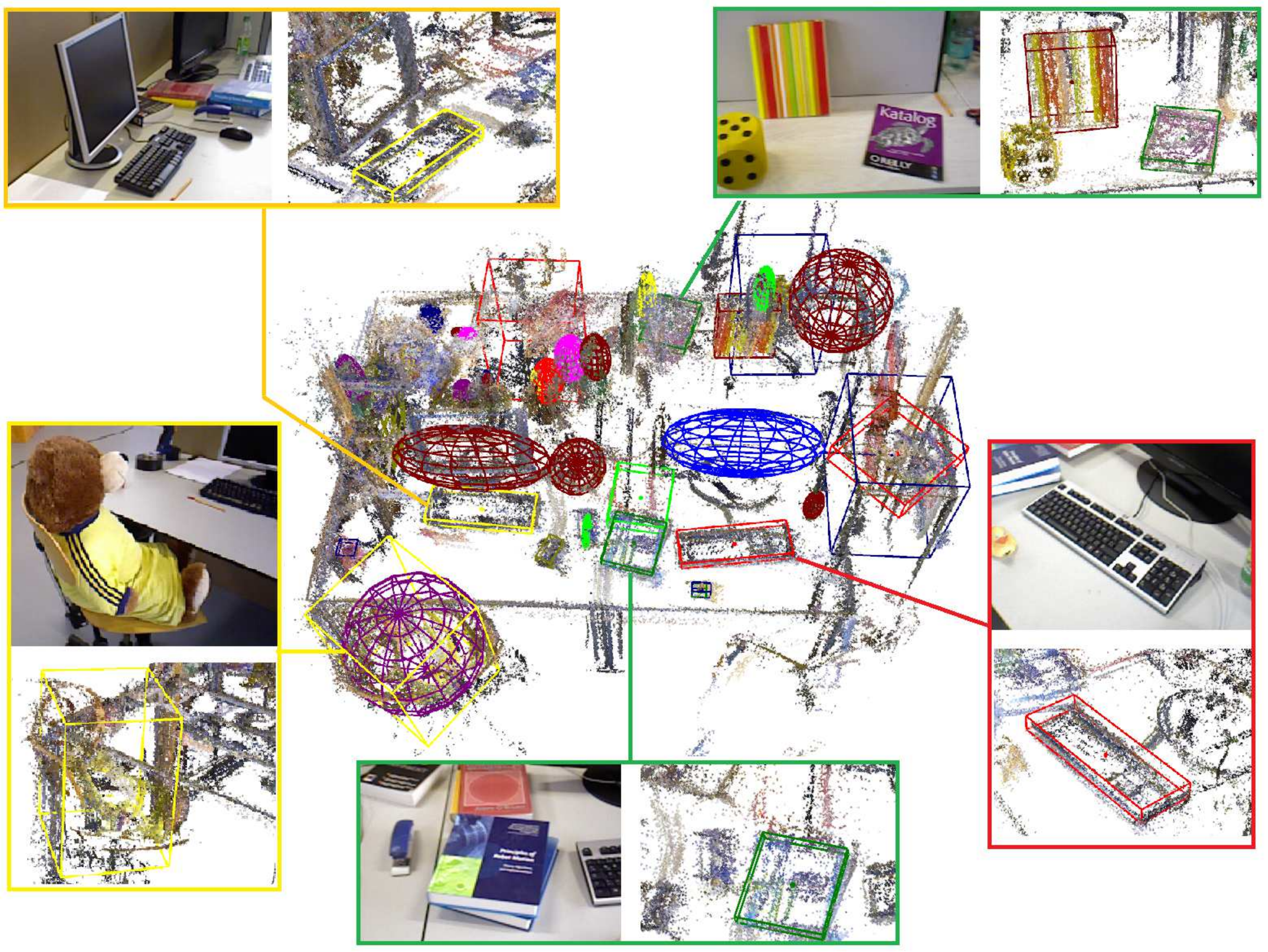}
	\caption{A lightweight and object-oriented semantic map.}
\label{Result}
\end{figure}

Object SLAM is a typical application of semantic SLAM, and the goal is to estimate more robust and accurate camera poses by leveraging the semantic information of in-frame objects \cite{13,14,li2017hybrid}. In this work, we further extend the content of object SLAM by enabling it to build lightweight and object-oriented maps, demonstrated in Fig. \ref{Result}, in which the objects are represented by cubes or quadrics with their locations, orientations, and scales accurately registered. 

The challenges of object SLAM mainly lie in two folds: 1) Existing data association methods \cite{3,20,5} are not robust or accurate for tackling complex environments that contain multiple object instances. There are no practical solutions to systematically address this problem.
2) Object pose estimation is not accurate, especially for monocular object SLAM. Although some improvements are achieved in recent studies \cite{2, 21,30}, they are typically dependent on strict assumptions, which are hard to fulfill in real-world applications.

In this paper, we propose the EAO-SLAM, a monocular object SLAM system, to address the data association and pose estimation problems. Firstly, we integrate the parametric and nonparametric statistic tests, and the traditional IoU-based method, to conduct model ensembling for data association. Compared with conventional methods, our approach sufficiently exploits the nature of different statistics, e.g., Gaussian and non-Gaussian measurements, hence exhibits significant advantages in association robustness. For object pose estimation, we propose a centroid and scale estimation algorithm and an object pose initialization approach based on the \textit{isolation forest} (iForest). The proposed methods are robust to outliers and exhibit high accuracy, which significantly facilitates the joint pose optimization process.

The contributions of this paper are summarized as follows:
\begin{itemize}
\item We propose an ensemble data association strategy that can effectively aggregate different measurements of the objects to improve association accuracy.  
\item We propose an object pose estimation framework based on iForest, which is robust to outliers and can accurately estimate the locations, poses, and scales of objects. 
\item Based on the proposed method, we implement the EAO-SALM to build lightweight and object-oriented maps.  
\item We conduct comprehensive experiments and verify the effectiveness of our proposed methods on publicly available datasets and the real scenario. The source code of this work is also released.  
\end{itemize}

\section{Related Work}
\subsection{Data Association}
Data association is an indispensable ingredient for semantic SLAM, which is used to determine whether the object observed in the current frame is an existing object in the map. Bowman \textit{et al.} \cite{3} use a probabilistic method to model the data association process and leverage the EM algorithm to find correspondences between observed landmarks. Subsequent studies \cite{5,6} further extend the idea to associate dynamic objects or conduct semantic dense reconstruction. These methods can achieve high association accuracy, but can only process a limited number of object instances. Their efficiency also remains to be improved due to the expensive EM optimization process \cite{9123604}.
Object tracking is another commonly-used approach in data association. Li \textit{et al.} \cite{11} propose to project 3D cubes to the image plane and then leverage the Hungarian tracking algorithm to conduct association using the projected 2D bounding boxes. Tracking-based methods perform high runtime efficiency, but can easily generate incorrect priors in complex environments, yielding incorrect association results.

In recent studies, more data association approaches are developed based on maximum shared information. Liu \textit{et al.} \cite{12} propose random walk descriptors to represent the topological relationships between objects, and those with the maximum number of shared descriptors are regarded as the same instance. Instead, Yang \textit{et al.} \cite{2} propose to directly count the number of matched map points on the detected objects as association criteria, yielding a much efficient performance. Grinvald \textit{et al.} \cite{13} propose to measure the similarity between semantic labels and Ok \textit{et al.} \cite{14} propose to leverage the correlation of hue saturation histogram. The major drawback of these methods is that the designed features or descriptors are typically not general or robust enough and can easily cause incorrect associations.

Weng \textit{et al.} \cite{19} for the first time propose nonparametric statistical testing for semantic data association, which can address the problems in which the statistics do not follow a Gaussian distribution. Later on, Iqbal \textit{et al.} \cite{20} also verify the effectiveness of nonparametric data association. However, this method cannot address the statistics that follow Gaussian distributions effectively, hence cannot sufficiently exploit different measurements in SLAM. Based on this observation, we combine the parametric and nonparametric methods to perform model ensembling, which exhibits superior association performance in the complex scenarios with the presence of multiple categories of objects.

\subsection{Object SLAM}
Benefiting from deep learning techniques \cite{zhu2018deep, zhu2018novel}, object detection is robustly integrated into the SLAM framework for labeling objects of interests in the map. The exploitation of in-frame objects significantly enlarges the application scopes of traditional SLAM. Some studies \cite{1,19,31} treat objects as landmarks to estimate camera poses or for  relocalization \cite{11}. Some studies \cite{10} leverage object size to constrain the scale of monocular SLAM, or remove dynamic objects to improve pose estimation accuracy \cite{5,16}. In recent years, the combination of object SLAM and grasping \cite{35} has also attracted many interests, and facilitate the research on autonomous mobile manipulation. 

Object models in semantic SLAM can be broadly divided into three categories: instance-level models, category-specific models, and general models. The instance-level models \cite{21, 22} depend on a well-established database that records all the related objects. The prior information of objects provides important object-camera constraints for graph optimization. Since the models need to be known in advance, the application scenarios of such methods are limited. There are also some studies on category-specific models, which focus on describing category-level features. For example, Parkhiya \textit{et al.} \cite{30} and Joshi \textit{et al.} \cite{31} use the CNN network to estimate the viewpoint of objects and then project the 3D line segments onto image planes to align them. The general model adopts simple geometric elements, e.g., cubes \cite{2,11}, quadrics \cite{1} and cylinders \cite{30}, to represent objects, which are also the most commonly-used models. 

In terms of the joint optimization of camera and object poses, Frost \textit{et al.} \cite{10} simply integrate object centroids as point clouds to the camera pose estimation process. Yang \textit{et al.} \cite{2} propose a joint camera-object-point optimization scheme to construct the pose and scale constraints for graph optimization. Nicholson \textit{et al.} \cite{1} propose to project the quadric onto the image plane and then calculates the scale error between the projected 2D rectangular and the detected bounding box. This work also adopts the joint optimization strategy, but with a novel initialization method, which can significantly improve the optimality of solutions. 

\section{System Overview}
\begin{figure}[t]
	\centering
	\captionsetup{belowskip=-10pt}
	\includegraphics[scale=0.35]{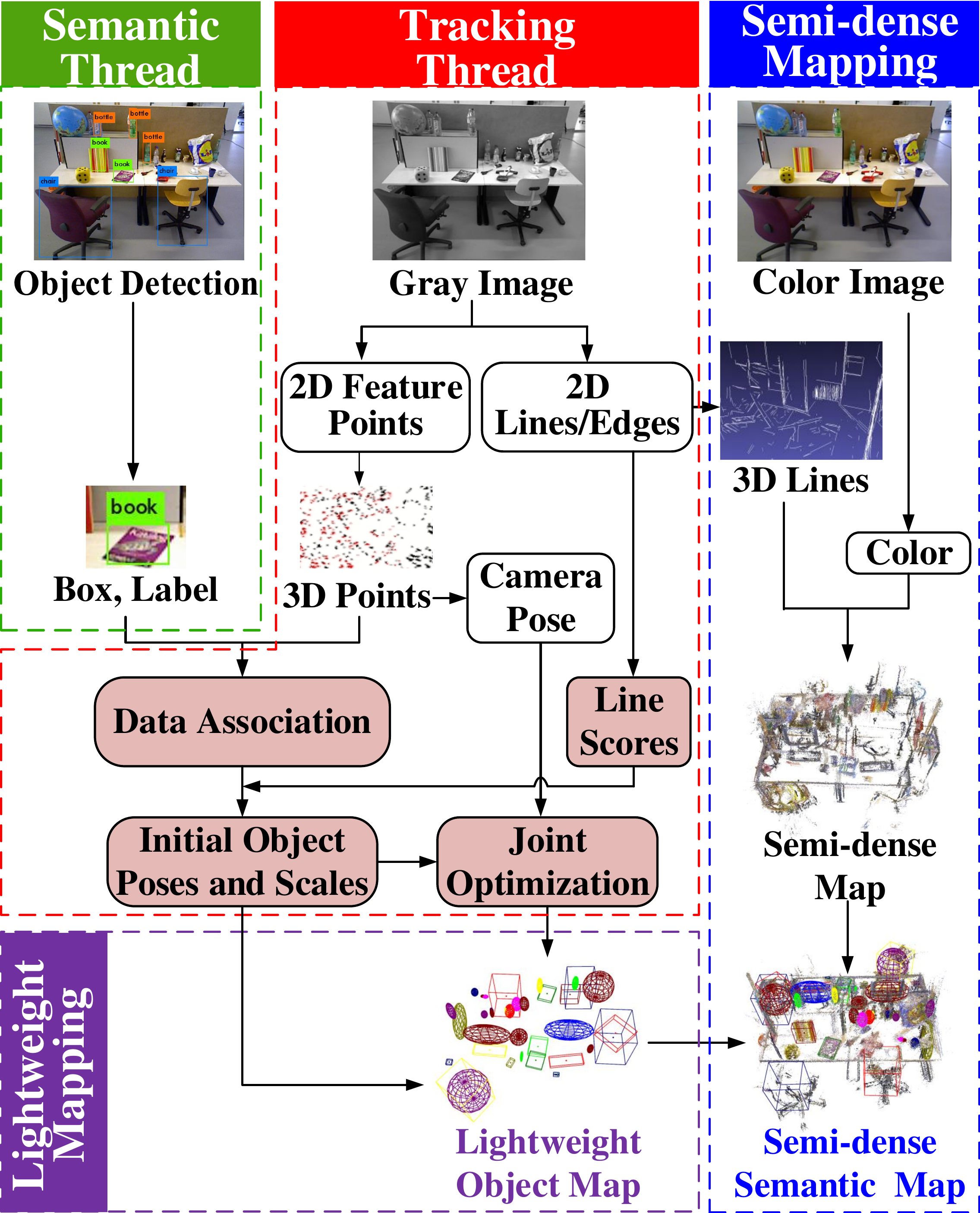}
	\caption{The architecture of EAO-SLAM system. The main contributions of this work are highlighted with red colors.}
	\label{Framework}
\end{figure}

The proposed object SLAM framework is demonstrated in Fig. \ref{Framework}, which is developed based on ORB-SLAM2 \cite{49}, and additionally integrates a semantic thread that adopts YOLOv3 as the object detector. The ensemble data association is implemented in the tracking thread, which combines the information of bounding boxes, semantic labels, and point clouds. After that, the iForest is leveraged to eliminate outliers for finding an accurate initialization for the joint optimization process. The object pose and scale are then optimized together with the camera pose to build a lightweight and object-oriented map. In semi-dense mapping thread, the object map is combined with a semi-dense map generated by \cite{48} to obtain the a semi-dense semantic map.

\section{Ensemble Data Association}

Throughout this section, the following notations are used:
\begin{itemize}
	\item $P \in \mathbb{R}^{3 \times |P|}, Q \in \mathbb{R}^{3 \times |Q|}$ - the point clouds of objects.
	\item $\mathcal{R}$ - the rank (position) of a data point in a sorted list.
	\item $\mathbf{c} \in \mathbb{R}^{3 \times 1}$ - the currently observed object centroid.
	\item $C=[\mathbf{c}_1, \mathbf{c}_2, \dots,  \mathbf{c}_{|C|}] \in \mathbb{R}^{3 \times |C|}$ - the history observations of the centroids of an object. $C_1,C_2$ are similar.
	\item $f(\cdot)$ - the probability function used for statistic test.
	\item $m(\cdot),\sigma(\cdot) \in \mathbb{R}^{3 \times 1}$ - the mean and variance functions.
\end{itemize} 


\subsection{Nonparametric Test}

\begin{figure}[t]
		\centering
	\captionsetup{belowskip=-10pt}
	\includegraphics[scale=0.28]{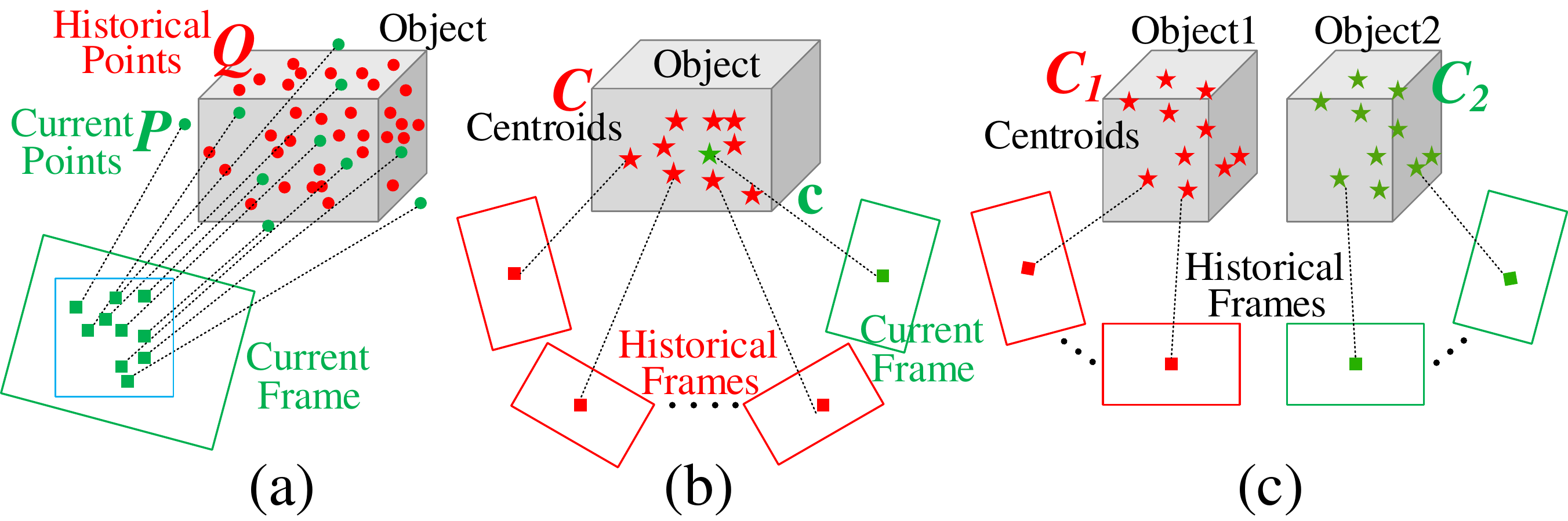}
	\caption{Different Types of Statistics Used for data association.}
	\label{DataAssociation}
\end{figure}

The nonparametric test is leveraged to process object point clouds (the red and green points in Fig. \ref{DataAssociation} (a)), which follows a non-Gaussian distribution according to our experimental studies (Section \ref{exp-es}). Theoretically, if $P$ and $Q$ belong to the same object, they should follow the same distribution, i.e., $f_P = f_Q$. We use the \textit{Wilcoxon Rank-Sum test} \cite{50} to verify whether the null hypothesis holds.

We first concatenate the two point clouds $X = [P|Q] = [\mathbf{x}_1, \mathbf{x}_2, \dots, \mathbf{x}_{|X|}]\in \mathbb{R}^{3\times(|P|+|Q|)}$, and then sort $X$ in three dimensions respectively. Define $W_{P} \in \mathbb{R}^{3\times1}$ as follows, 
\begin{equation}
\small
 W_{P}=\left \{ \sum_{k=1}^{|X|} \mathcal{R}(1\{\mathbf{x}_k\in P \})-\frac{|P|(|P|+1)}{2} \right \},
\vspace*{-0.4\baselineskip}
\end{equation}
and $W_{Q}$ is with the same formula. The Mann-Whitney statistics is $W \small{=} \min({W_{P}},{W_{Q}})$, which is proved to follow a Gaussian distribution asymptoticly \cite{50}. Herein, we essentially construct a Gaussian statistics using the non-Gaussian point clouds. The mean and variance of $W$ is calculated as follows:
\vspace*{-0.4\baselineskip}
\begin{align}
 m(W) &= (|P||Q|) / 2,\\
 \sigma(W) &= \frac{{|P||Q|\Delta}}{{12}} - \frac{{|P||Q|(\sum_{i} {\tau _i^3}  - \sum_{i} {{\tau _i}} )}}{{12(|P|+|Q|)\Delta}},
\vspace*{-0.4\baselineskip}
\end{align}
where $\Delta = |P|+|Q|+1$, and $\tau \in P \cap Q$. 
 
To make the null hypothesis stand, $W$ should meet the following constraints:
\vspace*{-0.4\baselineskip}
\begin{equation}
 \label{nonp}
 f(W) \geq f\left(r_r \right)=f\left(r_l \right)=\alpha/2,
\vspace*{-0.4\baselineskip}
\end{equation}
where $\alpha$ is the significance level, $1-\alpha$ is thus the confidence level, and $[r_l, r_r] \approx [m-s\sqrt{\sigma}, m+s\sqrt\sigma\,]$ defines the confidence region. The scalar $s > 0$ is defined on a normalized Gaussian distribution $\mathcal{N}(s|0,1) \small{=} \alpha$. In summary, if the Mann-Whitney statistics $W$ of two point clouds $P$ and $Q$ satisfies Eq. \eqref{nonp}, they come from the same object and the data association successes.

\subsection{Single-sample and Double-sample T-test}
The single-sample $t$-test is used to process object centroids observed in different frames (the stars in Fig. \ref{DataAssociation} (b)), which typically follow a Gaussian distribution (Section \ref{exp-es}). 

Suppose the null hypothesis is that $C$ and $\mathbf{c}$ are from the same object, and define $t$ statistics as follows,
\vspace*{-0.4\baselineskip}
\begin{equation}
t = \frac{m(C) - \boldsymbol{c}}{\sigma(C)/\sqrt{\left |C  \right |}}\sim t(\left |C  \right | - 1).
\vspace*{-0.4\baselineskip}
\end{equation}
To make the null hypothesis stand, $t$ should satisfy:
\vspace*{-0.4\baselineskip}
\begin{equation}
f(t)\geq f(t_{\alpha /2,v}) = \alpha /2
\label{condition of t-test}
\vspace*{-0.4\baselineskip}
\end{equation}
where $t_{\alpha /2,v}$ is the upper $\alpha /2$ quantile of the t-distribution of $v$ degrees of freedom, and $v=\sqrt{\left |C  \right |} - 1$. If $t$ statistics satisfies \eqref{condition of t-test}, $\mathbf{c}$ and $C$ comes from the same object.

Due to the strict data association strategy above or the bad angle of views, some existing objects may be recognized as new ones. Hence, a double-sample $t$-test is leveraged to determine whether to merge the two objects by testing their historical centroids (the stars in Fig. \ref{DataAssociation} (c)).

Construct $t$-statistics for $C_1$ and $C_2$ as follows,
\vspace*{-0.4\baselineskip}
\begin{equation}
t=\frac{m(C_1) - m(C_2)}{\sigma_d} \sim t(\left|C_1\right| + \left |C_2 \right | - 2)\\
\vspace*{-0.4\baselineskip}
\end{equation}
\vspace*{-0.4\baselineskip}
\begin{equation}
\small
\sigma_d = \sqrt {\frac{{\left( {{\left|C_1\right|} - 1} \right)\sigma_1^2 + \left( {{\left|C_2\right|} - 1} \right)\sigma_2^2}}{{{\left|C_1\right|} + {\left|C_2\right|} - 2}}\left( {\frac{1}{{{\left|C_1\right|}}} + \frac{1}{{{\left|C_2\right|}}}} \right)}
\vspace*{-0.4\baselineskip}
\end{equation}
where $\sigma_d$ is the pooled standard deviation of the two objects. Similarly, if $t$ satisfies \eqref{condition of t-test}, $v=\left|C_1\right| + \left |C_2 \right | - 2$, it means that $C_1$ and $C_2$ belongs to the same object, then we merge them.


\section{Object SLAM}
Throughout this section, the following notations are used:
\begin{itemize}
	\item $\boldsymbol{t}=[t_x, t_y, t_z]^T$ - the translation (location) of object frame in world frame.
		\item $\boldsymbol{\theta}=[\theta_r, \theta_y, \theta_p]^T$ - the rotation of object frame w.r.t. world frame. $R(\boldsymbol{\theta})$ is matrix representation.
		\item $T=\left \{ R(\boldsymbol{\theta}), \boldsymbol{t} \right \}$ - the transformation of object frame w.r.t. world frame.
		\item $\boldsymbol{s}=[s_l, s_w, s_h]^T$ - half of the side length of a 3D bounding box, i.e., the scale of an object.
		\item $P_o, P_w \in \mathbb{R}^{3 \times 8}$ - the coordinates of eight vertices of a cube in object and world frame, respectively. 
		\item $Q_o, Q_w \in \mathbb{R}^{4 \times 4}$ - the quadric parameterized by its semiaxis in object and world frame, respectively, where $Q_{o}=\operatorname{diag}\left\{s_{l}^{2}, s_{w}^{2}, s_{h}^{2},-1\right\}$. 
		\item $\alpha(\cdot)$ - calculate the angle of line segments. 
		\item $K,T_c$ - the intrinsic and extrinsic parameters of camera. 
		\item $\boldsymbol{p} \in \mathbb{R}^{3 \times 1}$ - the  coordinates of a point in world frame. 
\end{itemize} 

\begin{figure}[t]
		\centering
	\captionsetup{belowskip=-10pt}
	\includegraphics[scale=0.26]{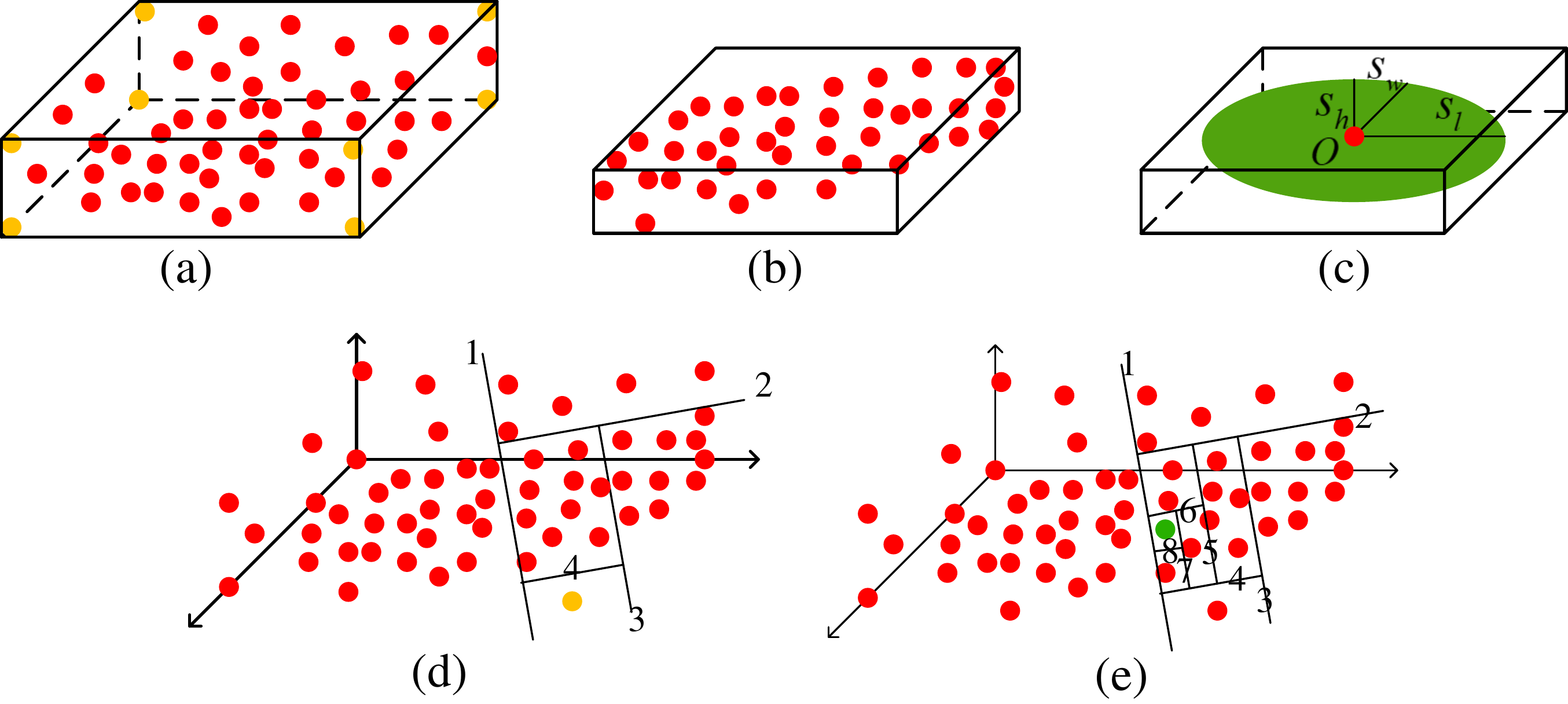}
	\caption{Object representation and demonstration of iForest.}
	\label{isolation}	
\end{figure}

\renewcommand{\algorithmicrequire}{\textbf{Input:}} 
\renewcommand{\algorithmicensure}{\textbf{Output:}}
\begin{algorithm}[t]
	\caption{Centroid and Scale Estimation Based on iForest}
	\label{algorithm1}
	\begin{algorithmic}[1] 
		\Require $X$ - The point cloud of an object, $t$ - The number of iTrees in iForest, $\psi$ - The subsampling size for an iTree.
		\Ensure $\mathcal{F}$ - The iForest, a set of iTrees, $\boldsymbol{t}$ - The origin of local frame, $\boldsymbol{s}$ - The initial scale of the object. 
		\vspace{1mm}		
		\Procedure{paraObject}{$X, t, \psi$}
		\State $\mathcal{F} \gets$ \Call{buildForest}{$X, t, \psi$}
		\For{point $\mathbf{x}$ in $X$}
		\State $E(h) \gets$ averageDepth($\mathbf{x},\mathcal{F}$) 
		\State $s \gets $ score($E(h), C$) \Comment Eq. \eqref{score} and \eqref{ptlen}		
		\If{$s > 0.6$} \Comment an empirical value
		\State remove($\mathbf{x}$) \Comment remove $\mathbf{x}$ from $X$
		\EndIf
		\EndFor
		\State $\boldsymbol{t} \gets$ meanValue($X$)
		\State $\boldsymbol{s} \gets$ (max($X$) - min($X$)) / 2
		\State \Return $\mathcal{F}, \boldsymbol{t}, \boldsymbol{s}$
		\EndProcedure
		\vspace{1mm}
		
		\Procedure{buildForest}{$X, t, \psi$}
		\State $\mathcal{F} \gets \phi$
		\State $l \gets $ ceiling($\log_2\psi$) \Comment maximum times of iterations
		\For{$i=$ 1 to $t$}
		\State $X^{ (i)}\gets$ randomSample($X, \psi$)
		\State $\mathcal{F} \gets \mathcal{F} \;\cup$ \Call{buildTree}{$X^{(i)}, 0, l$}
		\EndFor \\
		\Return $\mathcal{F}$
		\EndProcedure
		\vspace{1mm}
		
		\Procedure{buildTree}{$X, e, l$}
		\If{$e\geq l$ or $|X| \leq 1$}
		\State \Return exNode\{$|X|$\} \Comment record the size of $X$
		\EndIf
		\State $i \gets$ randomDim(1, 3) \Comment get one dimension
		\State $q \gets$ randomSpitPoint($X[i]$)
		\State $X_l, X_r \gets$ split($X[i], q$)
		\State $L\gets$\Call{buildTree}{$X_l, e+1, l$} \Comment get child pointer
		\State $R\gets$\Call{buildTree}{$X_r, e+1, l$}
		\State \Return inNode\{$L, R, i, q$\}
		\EndProcedure
	\end{algorithmic}
\end{algorithm}

	
\textbf{Object Representation}: In this work, we leverage the cubes and quadrics to represent objects, rather than the complex instance-level or category-level model. For objects with regular shapes, such as books, keyboards, and chairs, we use cubes (encoded by its vertices $P_o$) to represent them. For non-regular objects without an explicit direction, such as balls, bottles, and cups, the quadric (encoded by its semiaxis $Q_o$) is used for representation. Here, $P_o$ and $Q_o$ are expressed in object frame and only depend on the scale $\boldsymbol{s}$. To register these elements to global map, we also need to estimate their translation $\boldsymbol{t}$ and orientation $\boldsymbol{\theta}$ w.r.t. global frame. The cubes and quadrics in global frame are expressed as follows:
\begin{equation}
\label{cube para}
P_w= R(\boldsymbol{\theta})P_o + \boldsymbol{t},
\end{equation}
\begin{equation}
\label{quadric para}
Q_w= {T}Q_oT^T.
\end{equation}
With the assumption that the objects are placed parallel with the ground, i.e., $\theta_r\small{=}\theta_p\small{=}0$, we only need to estimate $[\theta_y, \boldsymbol{t}, \boldsymbol{s}]$ for a cube and $[\boldsymbol{t}, \boldsymbol{s}]$ for a quadric.

\vspace{1mm}
\textbf{Estimate $\boldsymbol{t}$ and $\boldsymbol{s}$}: Suppose there is an object point cloud $X$ in global frame, we follow conventions and denote its mean by $\boldsymbol{t}$, based on which, the scale can be calculated by $\boldsymbol{s}=(\max(X)-\min(X))/2$. The main challenge here is that $X$ is typically with many outliers, which can introduce a large bias to the estimation of $\boldsymbol{t}$ and $\boldsymbol{s}$. One of our major contributions in this paper is the development of an outlier-robust centroid and scale estimation algorithm based on the iForest \cite{47} to improve the estimation accuracy. The detailed procedure of our algorithm is presented in Alg. \ref{algorithm1}.

The key idea of the algorithm is to recursively separate the data space into a series of isolated data points, and then take the easily isolated ones as outliers. The philosophy is that, normal points is typically located more closely and thus need more steps to isolate, while the outliers usually scatter sparsely and can be easily isolated with less steps. 
As indicated by the algorithm, we first create $t$ isolated trees (the iForest) using the point cloud of an object (lines 2 and 14-33), and then identify the outliers by counting the path length of each point $\mathbf{x}\in X$ (lines 3-9), in which the score function is defined as follows:
\begin{equation}
\label{score}
\begin{split}
s(\mathbf{x})&=2\exp{\frac{-E(h(\mathbf{x}))}{C}},
\end{split}
\end{equation}
\begin{equation}
\label{ptlen}
\begin{split}
C=2 &H(|X|-1)-\frac{2(|X|-1)}{|X|},
\end{split}
\vspace*{-0.4\baselineskip}
\end{equation}
where $C$ is a normalization parameter, $H$ is a weight coefficient, and $h(\mathbf{x})$ is the height of point $\mathbf{x}$ in isolated tree. As demonstrated in Fig. \ref{isolation}(d)-(e), the yellow point is isolated after four steps, thus its path length is 4, and the green point has a path length of 8. Therefore, the yellow point is more likely to be an outlier.
In our implementation, points with a score greater than 0.6 are removed, and the remainings are used to calculate $\boldsymbol{t}$ and $\boldsymbol{s}$ (lines 10-12). Based on $\boldsymbol{s}$, we can initially construct the cubics and quadratics in the object frame, as shown in Fig. \ref{isolation}(a)-(c). $\boldsymbol{s}$ will be further optimized along with the object and camera poses later on. 

\begin{figure}[t]
	\centering
	\captionsetup{belowskip=-16pt}
	\includegraphics[scale=0.40]{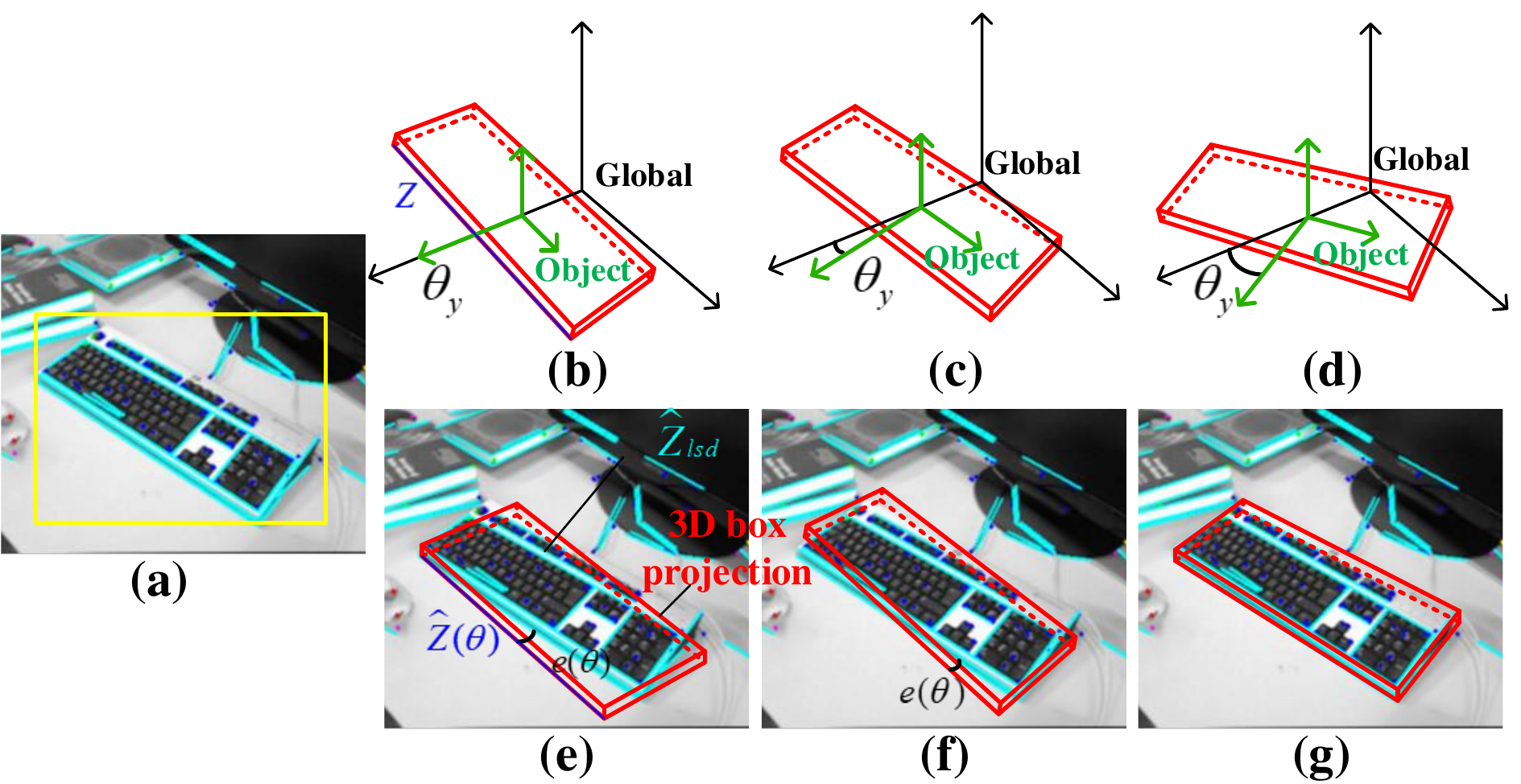}
	\caption{Line alignment to estimate object direction.}
	\label{SampleYaw}
\end{figure}

\vspace{1mm}
\textbf{Estimate $\theta_y$}: The estimation of $\theta_y$ is divided into two steps, namely to find a good initial value for $\theta_y$ first and then conduct numerical optimization based on the initial value. Since pose estimation is a non-linear process, a good initialization is very important to help improve the optimality of the estimation result. Conventional methods \cite{11} usually neglect the initialization process, which typically yields inaccurate results. 

The details of pose initialization algorithm is presented in Alg. \ref{algorithm2}. The inputs are obtained as follows: 1) LSD segments are extracted from $t$ consecutive image and those falling in the bounding boxes are assigned to the corresponding objects (see Fig. \ref{SampleYaw}a); 2) The initial pose of an object is assumed to be consistent with the global frame, i.e., $\theta_0\small{=}0$ (see Fig. \ref{SampleYaw}b). In the algorithm, we first uniformly sample thirty angles within $[-\pi/2, \pi/2]$ (line 2). For each sample, we then evaluate its score by calculating the accumulated angle errors between LSD segments $Z_{lsd}$ and the projected 2D edges of 3D edges $Z$ of the cube (lines 3-12). The error is defined as follows:  
\begin{equation}
\label{costf}
\begin{aligned}
&e(\boldsymbol{\theta})=||\alpha(\hat{Z}(\boldsymbol{\theta}))-\alpha(\hat{Z}_{lsd})||^2, \\
&\hat{Z}(\boldsymbol{\theta})=K T_{c}\left(R(\boldsymbol{\theta})Z+\boldsymbol{t}\right).\\
\end{aligned}
\end{equation}
A demonstration of the calculation of $e(\boldsymbol{\theta})$ is visualized in Fig. \ref{SampleYaw} (e)-(g).
The score function is defined as follows:
\begin{equation}
\label{cost}
\text { Score }=\frac{N_{\text {p}}}{N_{\text {a }}}(1+0.1(\xi- E(e))),
\end{equation}
where $N_{\text {a}}$ is the total number of line segments of the object in the current frame, $N_{\text {p}}$ is the number of line segments that satisfy $e < \xi$, $\xi$ is a manually defined error threshold (five degrees here), and $E(e)$ is the average error of these line segments with $e < \xi$.
After evaluating all the samples, we choose the one that achieves the highest score as the initial yaw angle for the following optimization process (line 13).
\begin{algorithm}[t]
	\caption{Initialization for Object Pose Estimation}
	\label{algorithm2}
	\begin{algorithmic}[1] 
		\Require $Z_1, Z_2, \dots, Z_t$ - Line segments detected by LSD in $t$ consecutive images, $\theta_0$ - The initial guess of yaw angel.	
		\Ensure  $\theta$ - The estimation result of yaw angel, $e$ - The estimation errors.
		\vspace{1mm}		
		\State $\mathcal{S}, \mathcal{E} \gets \phi$
		\State $ \mathbf{\Theta} \gets$ sampleAngles($\theta_0$, 30) \Comment see Fig. \ref{SampleYaw} (b)-(d)
		\For{sample $\theta$ in $\mathbf{\Theta}$} 
		\State $s_{\theta},e_{\theta} \gets 0$
		\For{$Z$ in \{$Z_1, Z_2, \dots, Z_t$\}}
		\State $s, e \gets$ score($\theta, Z$) \Comment Eq. \eqref{costf} and \eqref{cost}
		\State $s_{\theta} \gets s_{\theta} + s $ 
		\State $e_{\theta} \gets e_{\theta} + e $ 
		\EndFor
		\State $\mathcal{S} \gets \mathcal{S} \cup \{s_{\theta}\}$
		\State $\mathcal{E} \gets \mathcal{E} \cup \{e_{\theta}\}$		
		\EndFor	
		\State $\theta^* \gets$ argmax($\mathcal{S}$)
		\State \Return $\theta^*$, $e_{\theta^*}$
	\end{algorithmic}
\end{algorithm}


\vspace{1mm}
\textbf{Joint Optimization}: After obtaining the initial $\boldsymbol{S}$ and $\theta_y$, we then jointly optimize object and camera poses:
\begin{equation}
\small
\{O, T_c\}^{*}=\argminB_{\{\theta_y, \boldsymbol{s}\}} \sum\left( {e}(\boldsymbol{\theta}) + {e}(\boldsymbol{s}) \right)+\argminB_{\{T_c\}} \sum e(\boldsymbol{p}),
\end{equation}  
where the first term is the object pose error defined in Eq. \eqref{costf} and the scale error ${e}(\boldsymbol{s})$ defined as the distance between the projected edges of a cube and their nearest parallel LSD segments. The second term $e(\boldsymbol{p})$ is the commonly-sued reprojection error in traditional SLAM framework.

%

\section{Experimental Results}
\subsection{Distributions of Different Statistics}
\label{exp-es}

\begin{figure}[t]
	\centering
		\captionsetup{belowskip=-6pt}
	\includegraphics[scale=0.28]{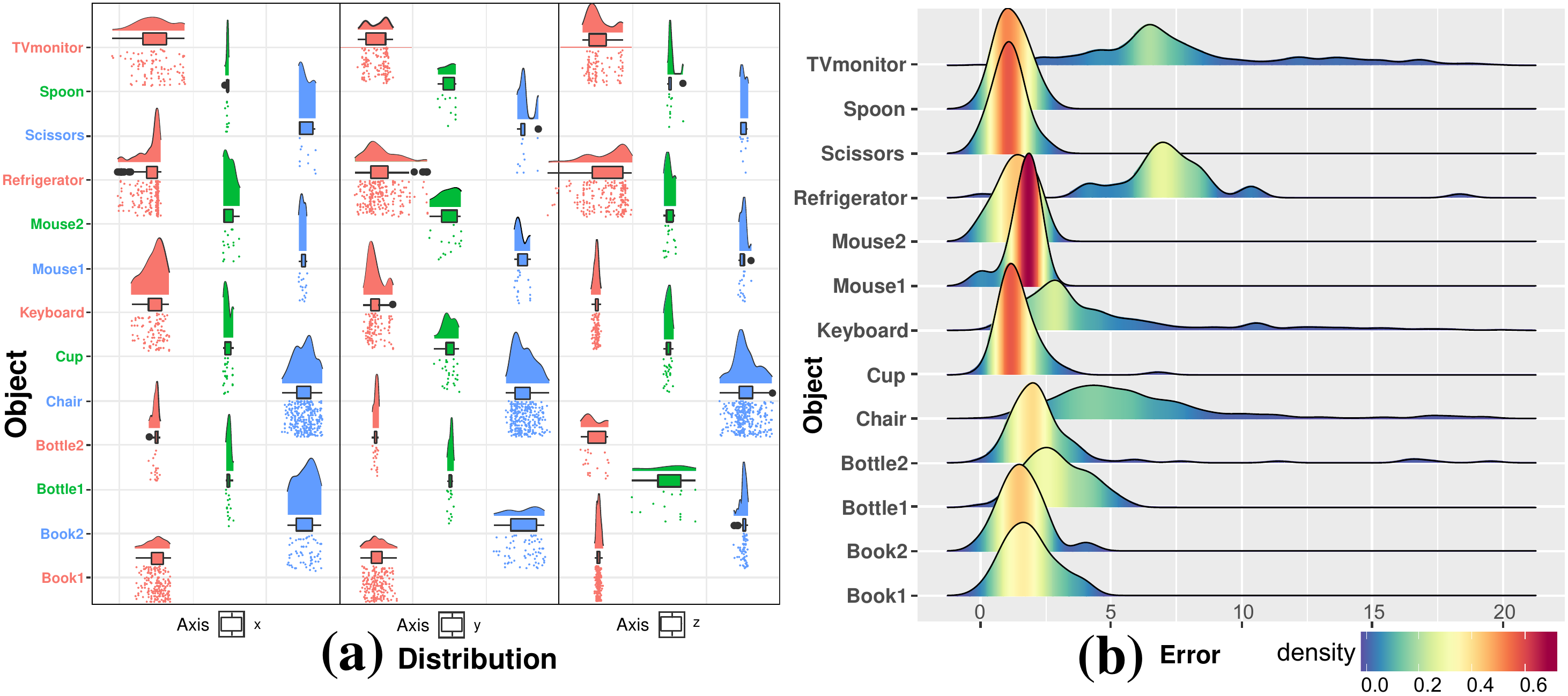}
	\caption{Distributions of different statistics in data association. (a) position distribution of point clouds in three directions. (b) distance error distribution of centroids.}
	\label{Distribution}
\end{figure}

For data association, the adopted statistics for statistical testing include the point clouds and their centroids of an object. To verify our hypothesis about the distributions of different statistics, we analyze a large amount of data and visualize their distributions in Fig. \ref{Distribution}.

Fig. \ref{Distribution} (a) shows the distributions of the point clouds of 13 objects during the data association in the fr3\_long\_office sequence. It is obvious that such statistics do not follow a Gaussian distribution. We can be seen that the distributions are related to specific characteristics of the objects, and do not show consistent behaviors. Fig. \ref{Distribution} (b) shows the error distribution of object centroids in different frames, which typically follow the Gaussian distribution. This result verifies the reasonability of applying the nonparametric \textit{Wilcoxon Rank-Sum test} for point clouds and t-test for object centroids.

\subsection{Ensemble Data Association Experiments}
\begin{figure}[t]
\centering
\includegraphics[scale=0.40]{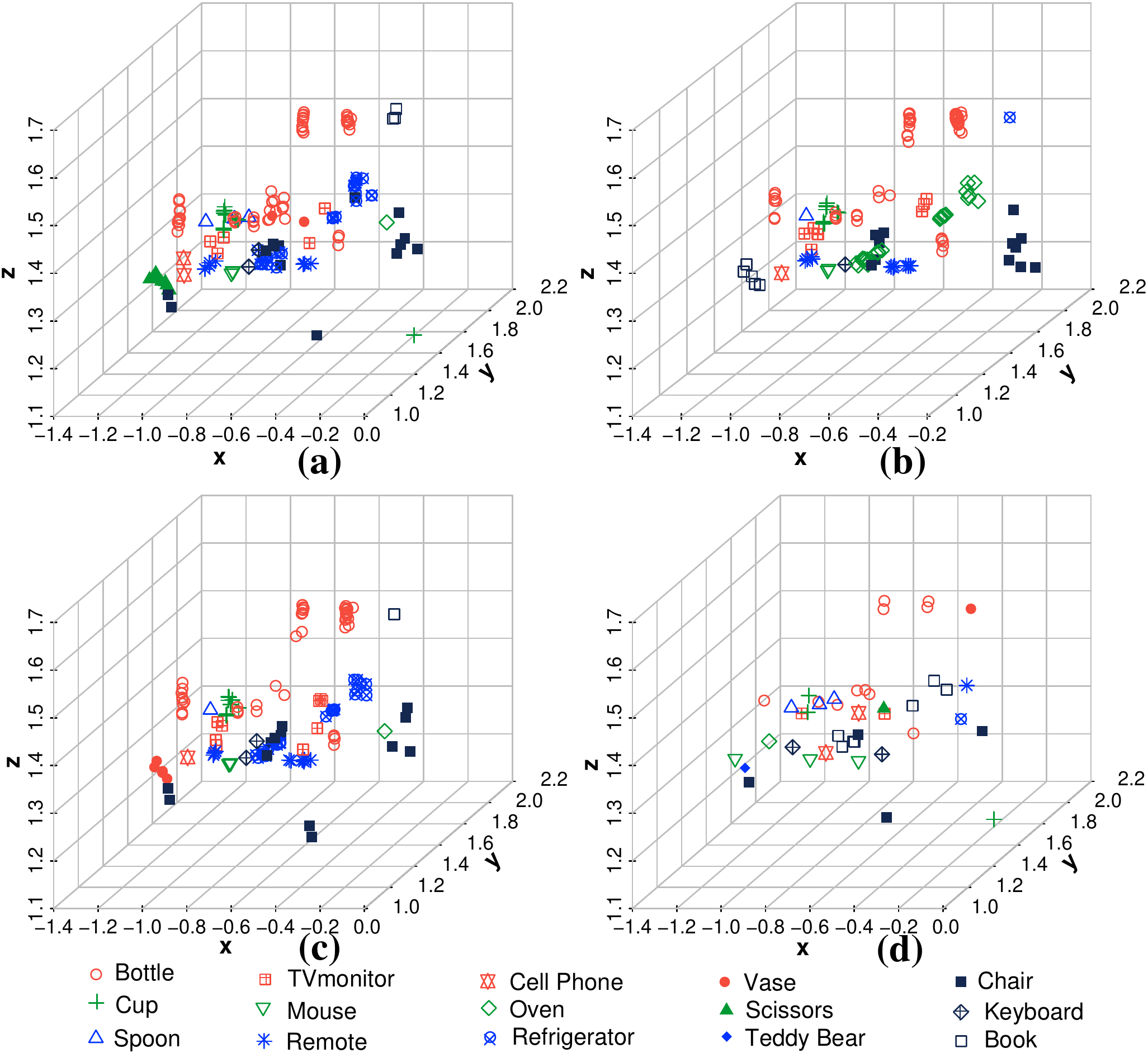}
\caption{Qualitative comparison of data association results. (a) IoU method. (b) IoU and nonparametric test. (c) IoU and t-test. (d) our ensemble method. }
\label{DataAssociationResults}
\end{figure}

We compare our method with the commonly-used Intersection over Union (IoU) method, nonparametric test (NP), and t-test. Fig. \ref{DataAssociationResults} shows the association results of these methods in TUM  fr3\_long\_office sequence. It can be seen that some objects are not correctly associated in (a)-(c). Due to the lack of association information, existing objects are often misrecognized as new ones by these methods once the objects are occluded or disappear in some frames, resulting in many unassociated objects in the map. In contrast, our method is much more robust and can effectively address this problem (see Fig. \ref{DataAssociationResults}(d)). The results of other sequences are shown in Table \ref{1}, and we use the same evaluation metric as \cite{20}, which measures the number of objects that finally present in the map. The \textit{GT} represents the ground-truth object number. As we can see, our method achieves a high success rate of association, and the number of objects in the map goes closer to GT, which significantly demonstrates the effectiveness of the proposed method.

We also compare our method with \cite{20}, and the results are shown in Table \ref{2}. As is indicated, our method can significantly outperform \cite{20}. Especially in the TUM dataset, the number of successfully associated objects by our method is almost twice than that by \cite{20}. In Microsoft RGBD and Scenes V2, the advantage is not obvious since the number of objects is limited there. Reasons of the inaccurate association of \cite{20} lie in two folds: 1) The method does not exploit different statistics and only used non-parametric statistics, thus resulting in many unassociated objects; 2) A clustering algorithm is leveraged to tackling the problem mentioned above, which removes most of the candidate objects.

\begin{table}[h]
		\renewcommand{\arraystretch}{1.2}
\setlength{\abovecaptionskip}{0.cm}
\setlength{\belowcaptionskip}{-0.1cm}
\caption{DATA ASSOCIATION RESULTS}
\begin{center}
\label{1}
\setlength{\tabcolsep}{3mm}{
\begin{tabular}{c|c|c|c|c|c}
\hline
& IoU & IoU$+$NP & IoU$+$t-test & EAO & GT\\
\hline
Fr1\_desk & 62 & 47 & 41 & {\bf 14} & 16\\
\hline
Fr2\_desk & 83 & 64 & 52 & {\bf 22} & 25\\
\hline
Fr3\_office & 150 & 128 & 130 & {\bf 42} & 45\\
\hline
Fr3\_teddy & 32 & 17 & 21 & {\bf 6} & 7\\
\hline
\end{tabular}}
\end{center}
\vspace*{-1.0\baselineskip}
\end{table}

\subsection{Qualitative Assessment of Object Pose Estimation}

\begin{figure}[!htbp]
	\centering
	\includegraphics[scale=0.42]{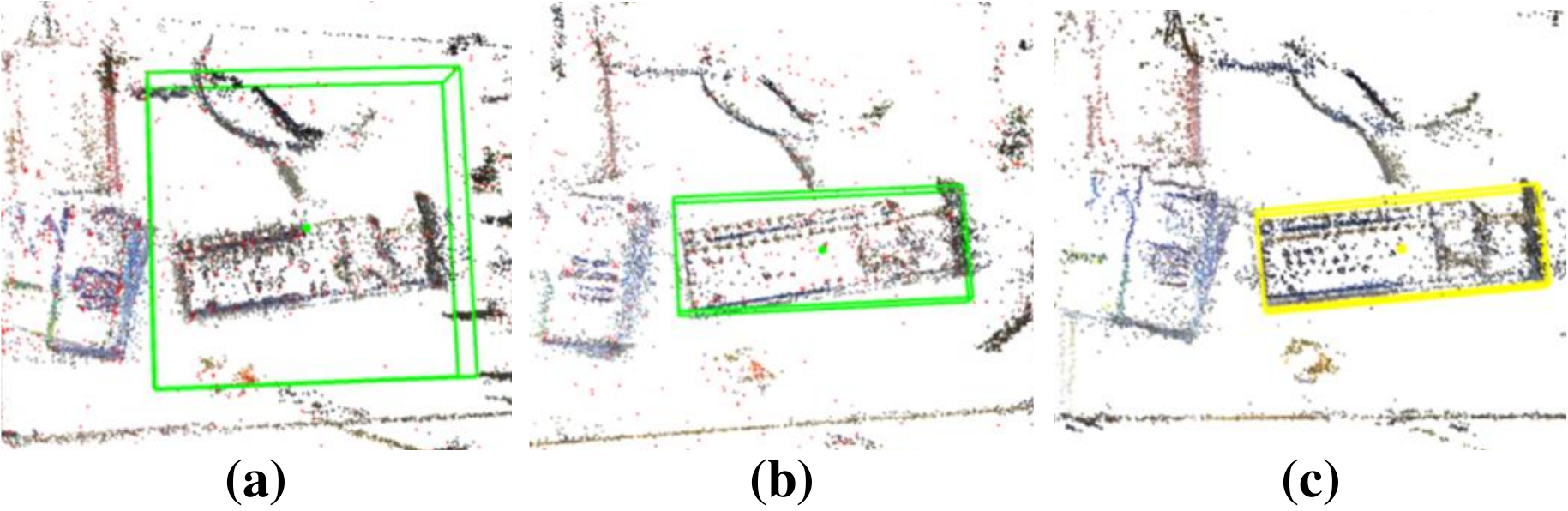}
	\caption{Visualization of the pose estimation.}
	\label{KeyboardResult}
\end{figure}
\begin{figure}[!htbp]
	\centering
	\captionsetup{belowskip=-10pt}
	\includegraphics[scale=0.10]{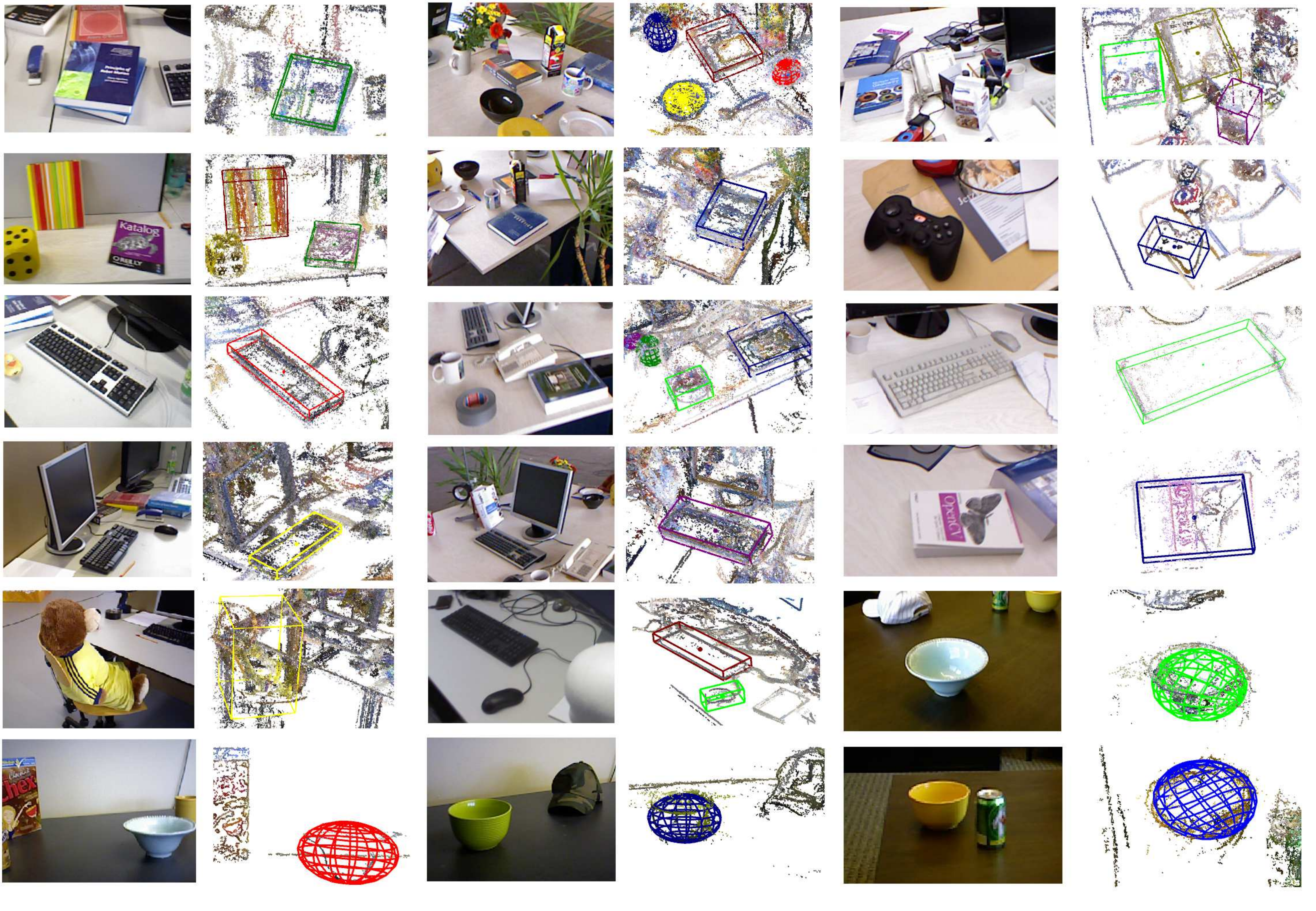}
	\caption{Results of object pose estimation. Odd columns: original RGB images. Even column: estimated object poses.}
	\label{ObjectPose}
\end{figure}

We superimpose the cubes and quadrics of objects on semi-dense maps for qualitative evaluation. Fig. \ref{KeyboardResult} is the 3D top view of a keyboard (Fig. \ref{SampleYaw}(a)) where the cube characterizes its pose. Fig. \ref{KeyboardResult}(a) is the initial pose with large scale error; Fig. \ref{KeyboardResult}(b) is the result after using iForest; Fig. \ref{KeyboardResult}(c) is  the final pose after our joint pose estimation. Fig. \ref{ObjectPose} presents the pose estimation results of the objects in 14 sequences of the three datasets, in which the objects are placed randomly and in different directions. As is shown, the proposed method achieves promising results with a monocular camera, which demonstrate the effectiveness of our pose estimation algorithm.
Since the datasets are not specially designed for object pose estimation, there is no ground truth for quantitatively evaluate the methods. 
Here, we compare $\theta_y$ before initialization (BI), after initialization (AI), and after joint optimization (JO). As shown in Table \ref{3}, the original direction of the object is parallel to the global frame, and there is a large angle error. After pose initialization, the error is decreased, and after the joint optimization, the error is further reduced, which verifies the effectiveness of our pose estimation algorithm.

\subsection{Object-Oriented Map Building}

\begin{figure*}[!htbp]
\centering
\includegraphics[scale=0.08]{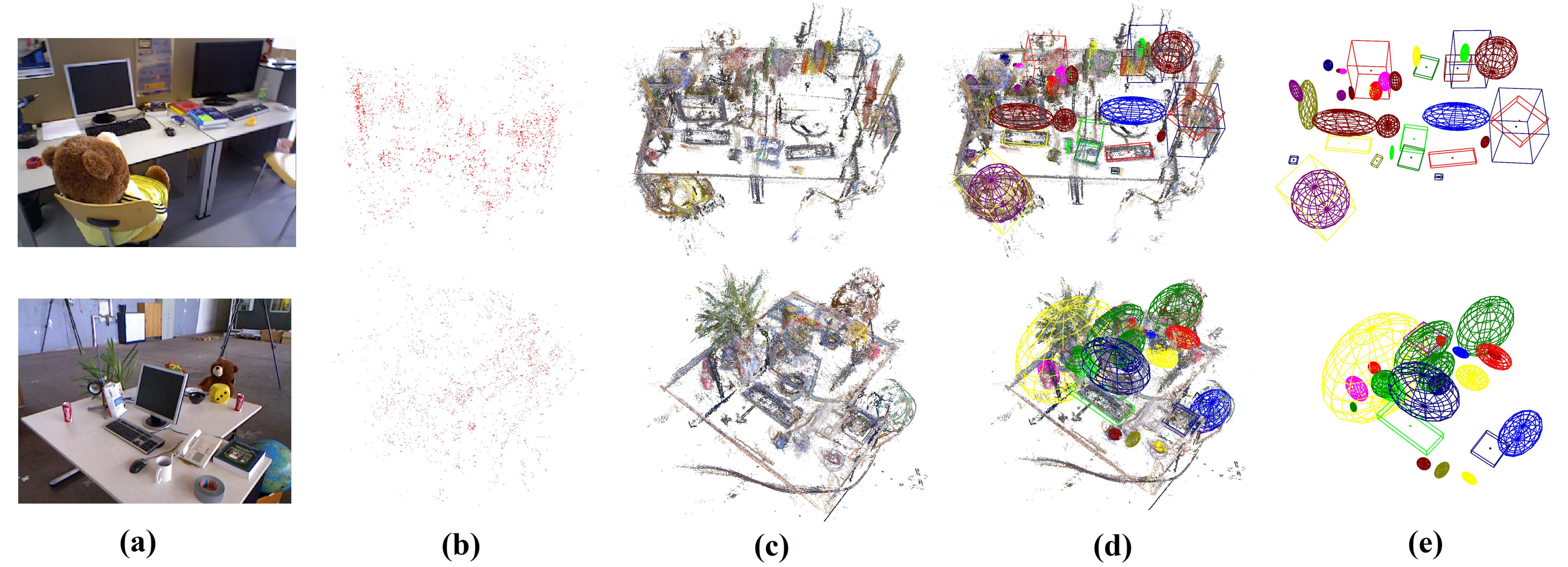}
\caption{Different map representations. (a) the RGB images. (b) the sparse map. (c) semi-dense map. (d) our semi-dense semantic map. (e) our lightweight and object-oriented map. (d) and (e) are build by the proposed EAO-SLAM.}
\label{TUM4Maps}
\end{figure*}

Lastly, we build the object-oriented semantic maps based on the robust data association algorithm, the accurate object pose estimation algorithm, and a semi-dense mapping system. Fig. \ref{TUM4Maps} shows two examples of TUM fr3\_long\_office and fr2\_desk, where (d) and (e) show a semi-dense semantic map and an object-oriented map, build by EAO-SLAM. Compared with the sparse map of ORB-SLAM2, our maps can express the environment much better. Moreover, the object-oriented map shows the superior performance in environment understanding than the semi-dense map proposed in \cite{48}.

\begin{figure*}[!htbp]
\centering
\captionsetup{belowskip=-10pt}
\includegraphics[scale=0.08]{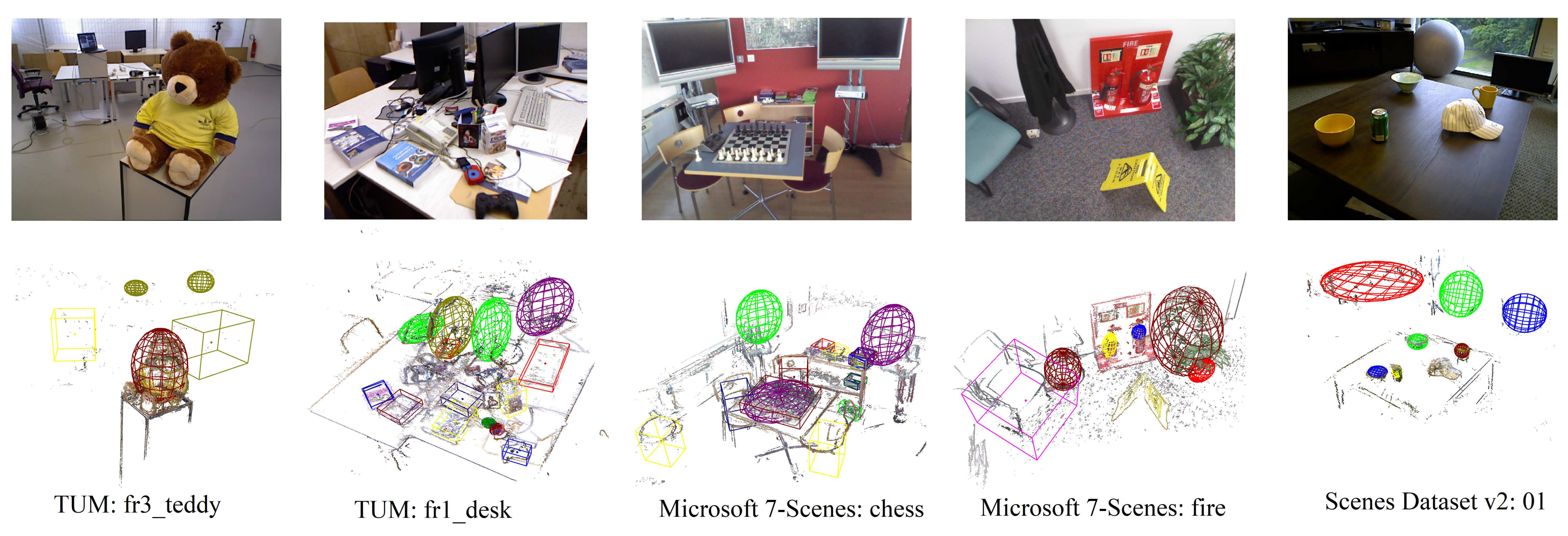}
\caption{Results of EAO-SLAM on the three datasets. Top: raw images. Bottom: simi-dense object-oriented map.}
\label{ThreeDatasetsResult}
\end{figure*}

\begin{figure*}[!htbp]
\centering
\captionsetup{belowskip=-10pt}
\includegraphics[scale=0.17]{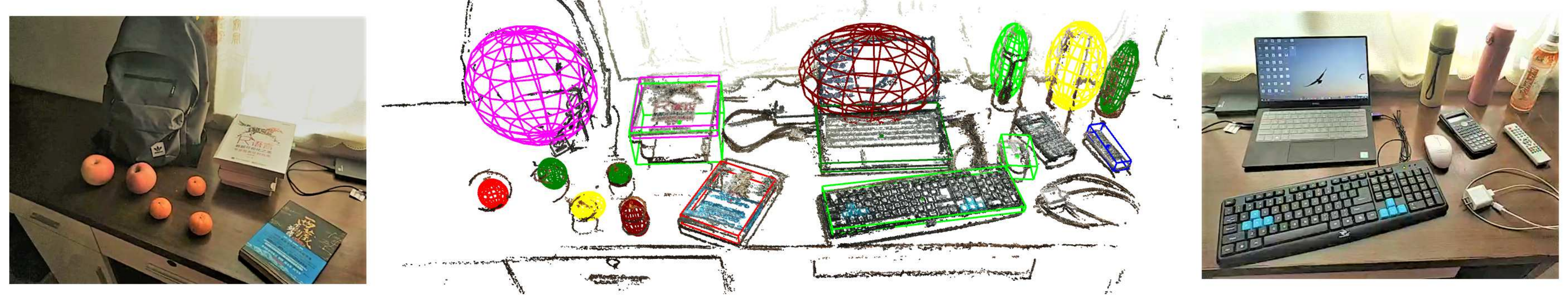}
\caption{Results of EAO-SLAM in a real scenario. Left and right: raw images. Middle: semi-dense object-oriented map.}
\label{RealScene}
\end{figure*}

The mapping results of other sequences in TUM, Microsoft RGB-D, and Scenes V2 datasets are shown in Fig. \ref{ThreeDatasetsResult}. It can be seen that EAO-SLAM can process multiple classes of objects with different scales and orientations in complex environments. Inevitably, there are some inaccurate estimations. For instance, in the \textit{fire} sequence, the chair is too large to be well observed by the fast moving camera, thus yielding an inaccurate estimation. We also conduct experiment in a real scenario, Fig. \ref{RealScene}. It can be seen even the objects are occluded, they can be accurately estimated, which further verifies the robustness and accuracy of our system.

%
%
%
%
%
%
%
%
%
%
%
%
%
%
%
%
%
%
%

\begin{table*}[!ht]
			\renewcommand{\arraystretch}{1.3}
	\caption{QUANTITATIVELY ANALYZED DATA ASSOCIATIONS}
	
	\begin{center}
		
		\label{2}
		\setlength{\tabcolsep}{2.1mm}{
			
			\begin{tabular}{c|c|c|c|c|c|c|c|c|c|c|c|c|c|c}
				
				\hline
				
				\multirow{2}*{Seq} & \multicolumn{4}{c|}{Tum} & \multicolumn{5}{c|}{Microsoft RGBD} & \multicolumn{5}{c}{Scenes V2}\\
				
				\cline{2-15}
				
				& fr1\_desk & fr2\_desk & fr3\_long\_office & fr3\_teddy & Chess & Fire & Office & Pumpkin & Heads & 01 & 07 & 10 & 13 & 14\\
				
				\hline
				
				\cite{20} & - & 11 & 15 & 2 & 5 & 4 & 10 & 4 & - & 5 & - & 6 & 3 & 4\\
				
				\hline
				
				{\bf Ours} & {\bf 14} & {\bf 22} & {\bf 42} & {\bf 6} & {\bf 13} & {\bf 6} & {\bf 21} & {\bf 6} & {\bf 15} & {\bf 7} & {\bf 7} & {\bf 7} & {\bf 3} & {\bf 5}\\
				
				\hline
				
				GT & 16 & 23 & 45 & 7 & 16 & 6 & 27 & 6 & 18 & 8 & 7 & 7 & 3 & 6\\
				
				\hline
				
			\end{tabular}}
		\end{center}
		
	\end{table*}

	\begin{table*}[!ht]
				\renewcommand{\arraystretch}{1.3}
		\caption{QUANTITATIVE ANALYSIS OF OBJECT ANGLE ERROR}
		
		\begin{center}
			
			\label{3}
			\setlength{\tabcolsep}{1.0mm}{
				
				\begin{tabular}{c|c|c|c|c|c|c|c|c|c|c|c|c|c|c|c}
					
					\hline
					
					Seq & \multicolumn{6}{c|}{fr3\_long\_office}	& \multicolumn{4}{c|}{fr1\_desk}	& \multicolumn{4}{c|}{fr2\_desk} &\multirow{2}*{Mean}\\
					
					\cline{1-15}
					
					Objects & book1 & book2 & book3 & keyboard1 & keyboard2 & mouse & Book1 & Book2 & Tvmonitor1 & Tvmonitor2 & keyboard & Book1 & Book2 & mouse & \\
					
					\hline
					
					BI & 19.2 & 11.4 & 16.2 & 10.3 & 7.4 & 11.3 & 33.5 & 15.2 & 32.7 & 22.5 & 8.9 & 15.5 & 16.9 & 8.7 & 16.4\\
					
					
					
					\hline
					
					AI & 5.3 & 5.5 & 6.2 & 7.2 & 4.2 & 6.4 & 8.6 & 8.9 & {\bf 6.0} & 11.4 & 5.5 & {\bf 3.8} & 10.1 & {\bf 7.5} & 6.9\\
					
					
					
					\hline
					
					JO & {\bf 3.1} & {\bf 4.3} & {\bf 5.7} & {\bf 2.5} & {\bf 2.8} & {\bf 4.3} & {\bf 5.4} & {\bf 7.6} & 8.7 & {\bf 10.2} & {\bf 3.9} & 5.1 & {\bf 6.4} & 7.9 & {\bf 5.6}\\
					
					
					
					\hline
					
				\end{tabular}}
			\end{center}
			
		\end{table*}


\section{Conclusion}
In this paper, we present the EAO-SLAM system that aims to build semi-dense or lightweight object-oriented maps. The system is implemented based on a robust ensemble data association method and an accurate pose estimation framework. Extensive experiments show that our proposed algorithms and SLAM system can build accurate object-oriented maps with object poses and scales accurately registered. The methodologies presented in this work further push the limits of semantic SLAM and will facilitate related researches on robot navigation, mobile manipulation, and human-robot interaction.

%
%
%
%

\bibliographystyle{IEEEtran}
\bibliography{MPSLAM}\

\end{document}